\documentclass[a4paper,11pt]{article}

\usepackage{algorithm}
\usepackage{algpseudocode}

\usepackage{amsmath,amssymb}
\usepackage{bm}

\usepackage{booktabs}

\usepackage[usenames,dvipsnames,svgnames]{xcolor}
\usepackage{tango-colors}

\usepackage{tikz}
\usetikzlibrary{arrows}
\usetikzlibrary{shapes}
\usetikzlibrary{patterns}
\usepackage{tikz-qtree}
\tikzstyle{edge}=[->, >=stealth', shorten <=1pt, shorten >=1pt, auto, semithick]

\usepackage[a4paper, left=3cm, right=2.5cm, top=3cm, bottom=3cm]{geometry}

\usepackage{hyperref}

\usepackage{authblk}
\usepackage{natbib}
\usepackage{doi}

\usepackage[utf8]{inputenc}
\usepackage[T1]{fontenc}
\usepackage[english]{babel}
\usepackage{csquotes}

\renewcommand{\cite}{\citep}

\newcommand{\R}{\mathbb{R}}
\newcommand{\effic}{\mathcal{O}}

\newcommand{\grad}{\nabla}
\newcommand{\mat}[1]{\bm{#1}}
\DeclareMathOperator*{\argmax}{argmax}

\newcommand{\dataset}{\mathcal{X}}

\newcommand{\dims}{n}

\newcommand{\fun}{f}

\newcommand{\lbl}{\ell}
\newcommand{\nlabels}{L}

\newcommand{\dist}{d}
\newcommand{\kernel}{k}

\newcommand{\alphabet}{\mathcal{A}}
\newcommand{\sym}[1]{\text{#1}}
\newcommand{\trees}{\mathcal{T}}
\newcommand{\tree}{T}

\newcommand{\edit}{\delta}

\newcommand{\rep}{\mathrm{rep}}
\newcommand{\del}{\mathrm{del}}
\newcommand{\ins}{\mathrm{ins}}
\newcommand{\script}{\bar \delta}
\newcommand{\editidx}{j}
\newcommand{\editlim}{n}

\newbool{arxiv}
\setbool{arxiv}{true}

\author{Benjamin Paaßen}
\affil{CITEC Center of Excellence, Bielefeld University\thanks{Support by the Bielefeld Young Researchers Fund is gratefully acknowledged.}}

\title{Adversarial Edit Attacks for Tree Data}

\date{Preprint of the IDEAL 2019 paper \citet{Paassen2019IDEAL} as provided by the authors.}

\begin{document}

\maketitle

\pagestyle{myheadings}
\markright{Preprint of \citet{Paassen2019IDEAL} provided by the authors.}

\begin{abstract}
Many machine learning models can be attacked with adversarial examples,
i.e.\ inputs close to correctly classified examples that are classified incorrectly.
However, most research on adversarial attacks to date is limited to vectorial data,
in particular image data. In this contribution, we extend the field by introducing
adversarial edit attacks for tree-structured data with potential applications in
medicine and automated program analysis.
Our approach solely relies on the tree edit distance and a logarithmic number of
black-box queries to the attacked classifier without any need for gradient information.

We evaluate our approach on two programming and two biomedical data sets and show
that many established tree classifiers, like tree-kernel-SVMs and recursive neural
networks, can be attacked effectively.

\end{abstract}

\section{Introduction}

In recent years, multiple papers have demonstrated that machine learning classifiers can be fooled
by \emph{adversarial examples}, i.e.\ an example $x'$ that is close to a correctly classified
data point $x$, but is classified incorrectly \cite{Akhtar2018,Madry2018}.
The threat of such attacks is not to be underestimated, especially in security-critical
applications such as medicine or autonomous driving, where adversarial examples could lead
to misdiagnoses or crashes \cite{Eykholt2018}.

Despite this serious threat to \emph{all} classification models, existing
research has almost exclusively focused on image data \cite{Akhtar2018,Madry2018},
with the notable exceptions of a few contributions on 
audio data \cite{Carlini2018}, text data \cite{Ebrahimi2018}, and graph data \cite{Dai2018,Zuegner2018}.
In particular, no adversarial attack approach has yet been developed for tree data, such as
syntax trees of computer programs or biomedical molecules. Furthermore, all attack approaches
for non-image data to date rely on knowledge about the classifier architecture and/or gradient, which
may not always be available \cite{Madry2018}.

In this paper, we address both issues by introducing adversarial edit attacks, a novel black-box
attack scheme for tree data. In particular, we propose to select for a point $x$ a neighboring
point with a different label $y$, compute the tree edits necessary to change $x$ into $y$,
and applying the minimum number of edits which still change the classifier output.

Our paper is structured as follows. We first introduce background and related work on
adversarial examples, then introduce our adversarial attack method,
and finally evaluate our method by attacking seven different tree classifiers on four tree
data sets, two from the programming domain and two from the biomedical domain.

\section{Related Work}

Following Szegedy et al.\ \cite{Szegedy2014}, we define an adversarial example for some data
point $x \in \dataset$ and a classifier $\fun : \dataset \to \{1, \ldots, \nlabels \}$ and
a target label $\lbl \in  \{1, \ldots, \nlabels\}$ as the solution $z$ to the following
optimization problem
\begin{equation}
\min_{z \in \dataset, \text{s.t.} \fun(z) = \lbl} \quad \dist(z, x)^2,
\label{eq:adv}
\end{equation}
where $\dist$ is a distance on the data space $\dataset$.
In other words, $z$ is the closest data point to $x$ which is still classified as $\lbl$.
For image data, the distance $\dist(z, x)$ is often so small that $z$ and $x$ look exactly
the same to human observers \cite{Szegedy2014}.

Note that Problem~\ref{eq:adv} is hard to solve because $\dataset$ is typically high dimensional
and the constraint $\fun(z) = \lbl$ is discrete. Accordingly, the problem has been addressed with heuristic
approaches, such as the fast gradient sign method \cite{Goodfellow2014},
which changes $x$ along the sign of the gradient of the classifier loss; or Carlini-Wagner attacks, which incorporate
the discrete label constraint as a differentiable term in the objective function
\cite{Carlini2017}. We call these methods \emph{white-box} because they all rely on knowledge
of the architecture and/or gradient $\grad_z \fun(z)$ of the classifier. In contrast, there
also exist \emph{black-box} attack methods, which only need to query $\fun$ itself,
such as one-pixel attacks, which are based on evolutionary optimization instead of
gradient-based optimization \cite{Akhtar2018,Su2017}.

In the realm of non-image data, prior research has exclusively focused on white-box attacks for specific
data types and/or models. In particular, \cite{Carlini2018} consider audio files, relying on decibels
and the CTC loss as measure of distance; \cite{Ebrahimi2018} attack text data by inferring
single character replacements that increase the classification loss; and \cite{Dai2018,Zuegner2018}
attack graph data by inferring edge deletions or insertions which fool a graph convolutional
neural network model.

Our own approach is related to \cite{Carlini2018}, in that we rely on an alignment between two
inputs to construct adversarial examples, and to \cite{Ebrahimi2018}, in that we consider discrete
node-level changes, i.e.\ node deletions, replacements, or insertions. However, in contrast to
these prior works, our approach is black-box instead of white-box and works in tree data as well
as sequence data.

\section{Method}

To develop an adversarial attack scheme for tree data, we face two challenges. First,
Problem~\ref{eq:adv} requires a distance function $\dist$ for trees.
Second, we need a method to apply small changes to a tree $x$ in order to construct an
adversarial tree $z$.
We can address both challenges with the \emph{tree edit distance}, which is defined as
the minimum number of node deletions, replacements, or insertions needed to change a
tree into another \cite{Zhang1989} and thus provides both a distance and a change model.

Formally, we define a tree over some finite alphabet $\alphabet$ recursively as an expression
$\tree = x(\tree_1, \ldots, \tree_m)$, where $x \in \alphabet$ and where
$\tree_1, \ldots, \tree_m$ is a (possibly empty) list of trees over $\alphabet$.
We denote the set of all trees over $\alphabet$ as $\trees(\alphabet)$. As an example,
$\sym{a}()$, $\sym{a(b)}$, and $\sym{a(b(a, a), a)}$ are both trees over the alphabet
$\alphabet = \{\sym{a, b}\}$. We define the \emph{size} of a tree $\tree = x(\tree_1, \ldots, \tree_m)$
recursively as $|\tree| := 1 + \sum_{c=1}^m |\tree_c|$.

\begin{figure}
\begin{center}
\begin{tikzpicture}[level distance=0.8cm]
\Tree [.{$\sym{a}$}
[.\node (b) {$\sym{b}$}; {$\sym{c}$} {$\sym{d}$} ]
{$\sym{e}$}
]

\begin{scope}[shift={(+3,0)}]
\Tree [.{$\sym{a}$}
\node (c) {$\sym{c}$};
{$\sym{d}$}
\node (e) {$\sym{e}$};
]
\end{scope}

\path[edge]%
(b) edge [bend left] node[above] {$\del_{\rightarrow 2}$} (c);

\begin{scope}[shift={(+6,0)}]
\Tree [.{$\sym{a}$}
{$\sym{c}$}
\node (d) {$\sym{d}$};
\node (f) {$\sym{f}$};
]
\end{scope}

\path[edge]%
(e) edge [bend left] node[above] {$\rep_{\sym{f} \rightarrow 4}$} (f);

\begin{scope}[shift={(+9,0)}]
\Tree [.$\sym{a}$
$\sym{f}$
[.\node (g) {$\sym{g}$}; $\sym{d}$ ]
$\sym{e}$
]
\end{scope}

\path[edge]%
(d) edge [bend left] node[above] {$\ins_{2, 1, \sym{g} \rightarrow 1}$} (g);
\end{tikzpicture}
\end{center}
\caption{An illustration of the effect of the tree edit script $\script = \del_2,
\rep_{4, \sym{f}}, \ins_{1, 2, 1, \sym{g}}$ on the tree $\sym{(a(b(c,d), e))}$.
We first delete the second node of the tree, then replace the fourth node with an $\sym{f}$,
and finally insert a $\sym{g}$ as second child of the first node, using the former second child
as grandchild.}
\label{fig:edits}
\end{figure}
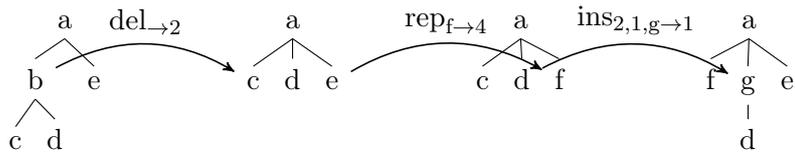

Next, we define a \emph{tree edit} $\edit$ over alphabet $\alphabet$ as a function $\edit : \trees(\alphabet)
\to \trees(\alphabet)$. In more detail, we consider node deletions $\del_i$, replacements $\rep_{i, a}$,
and insertions $\ins_{i, c, C, a}$, which respectively delete the $i$th node in the input tree
and move its children up to the parent, relabel the $i$th node in the input tree with symbol $a \in \alphabet$,
and insert a new node with label $a$ as $c$th child of node $i$, moving former children $c, \ldots, c + C$
down. Figure~\ref{fig:edits} displays the effects of each edit type.

We define an \emph{edit script} as a sequence
$\script = \edit_1, \ldots, \edit_\editlim$ of tree edits $\edit_\editidx$ and
we define the application of $\script$ to a tree $\tree$ recursively as
$\script(\tree) := (\edit_2, \ldots, \edit_\editlim)\big( \edit_1(\tree) \big)$.
Figure~\ref{fig:edits} displays an example edit script.

Finally, we define the \emph{tree edit distance} $\dist(x, y)$ as the length of the shortest script
which transforms $x$ into $y$, i.e.\ $\dist(x, y) := \min_{\script : \script(x) = y} |\script|$. This tree edit distance can
be computed efficiently via dynamic programming in $\effic(|x|^2 \cdot |y|^2)$ \cite{Zhang1989}.
We note that several variations of the tree edit distance with other edit
models exist, which are readily compatible with our approach \cite{Bille2005,Paassen2018ICML}.
For brevity, we focus on the classic tree edit distance in this paper.

\paragraph{Random baseline attack:} The concept of tree edits yields a
baseline attack approach for trees. Starting from a tree $x$ with label $\fun(x)$, we apply
random tree edits, yielding another tree $z$, until $\fun(z) \neq \fun(x)$. To make this more
efficient, we double the number of edits in each iteration until
$\fun(z) \neq \fun(x)$, yielding an edit script $\script = \edit_1, \ldots, \edit_\editlim$,
and then use binary search to identify the shortest prefix $\script_\editidx := \edit_1, \ldots, \edit_\editidx$
such that $\fun\big((\edit_1, \ldots, \edit_\editidx)(x)\big) \neq \fun(x)$.
This reduced the number of queries to $\effic(\log(\editlim))$.

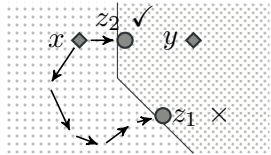
\begin{figure}
\begin{center}
\begin{tikzpicture}[node distance=0.3cm]
\draw[pattern=dots, pattern color=aluminium3, draw=none] (0.5,-0.5) -- (2,-0.5) -- (2,-1.5) -- (3,-2.5) -- (0.5,-2.5) -- cycle;
\draw[pattern=crosshatch dots, pattern color=aluminium3, draw=none] (2,-0.5) -- (2,-1.5) -- (3,-2.5) -- (4,-2.5) -- (4,-0.5) -- cycle;
\draw[draw=aluminium6] (2,-0.5) -- (2,-1.5) -- (3,-2.5);

\node[fill=aluminium4,draw=aluminium6, thick, minimum size=0.2cm, inner sep=0pt, diamond] (x) at (1.5,-1) {};
\node [left of=x] {$x$};

\node[fill=aluminium4,draw=aluminium6, thick, minimum size=0.2cm, inner sep=0pt, diamond] (y) at (3,-1) {};
\node [left of=y] {$y$};

\node[fill=aluminium4,draw=aluminium6, thick, minimum size=0.2cm, inner sep=0pt, circle] (z1) at (2.6,-2) {};
\node[right of=z1, node distance=0.5cm] {$z_1$ $\times$};

\node[fill=aluminium4,draw=aluminium6, thick, minimum size=0.2cm, inner sep=0pt, circle] (z2) at (2.1,-1) {};
\node[above of=z2] {$z_2$ $\checkmark$};

\node[inner sep=0pt] (xz1) at (1.1,-1.6) {};
\node[inner sep=0pt] (xz2) at (1.4,-2.25) {};
\node[inner sep=0pt] (xz3) at (1.8,-2.4) {};
\node[inner sep=0pt] (xz4) at (2.2,-2.1) {};

\path[edge]
(x) edge (xz1)
(xz1) edge (xz2)
(xz2) edge (xz3)
(xz3) edge (xz4)
(xz4) edge (z1);

\path[edge] (x) edge (z2);

\end{tikzpicture}
\end{center}
\caption{Two adversarial attack attempts, one random ($z_1$)
and one backtracing attack ($z_2$). $z_1$ is constructed by moving randomly in the
space of possible trees until the label changes. $z_2$ is constructed by moving
along the connecting line to the closest neighbor with different label $y$ until
the label changes. $z_1$ is \emph{not} counted as successful, because it is closer
to $y$ than to $x$, whereas $z_2$ is counted as successful.
The background pattern indicates the predicted label of the classifier.}
\label{fig:adv}
\end{figure}

Note that this random attack scheme may find solutions $z$ which are far away from $x$, thus
limiting the plausibility as adversarial examples. To account for such cases, we restrict
Problem~\ref{eq:adv} further and impose that $z$ only counts as a solution if $z$ is still closer to
$x$ than to any point $y$ which is correctly classified and has a different label than $x$ (refer to Figure~\ref{fig:adv}).

Another drawback of our random baseline is that it can not guarantee results after a fixed
amount of edits because we may not yet have explored enough trees to have crossed the classification
boundary. We address this limitation with our proposed attack method, \emph{backtracing attacks}.

\paragraph{Backtracing attack:} For any two trees $x$ and $y$, we can compute a
co-optimal edit script $\script$ with $\script(x) = y$ and $|\script| = \dist(x, y)$ in
$\effic(|x| \cdot |y| \cdot (|x| + |y|))$ via a technique called \emph{backtracing}
\cite[refer to Algorithm 6 and Theorem 16]{Paassen2018arxiv}.
This forms the basis for our proposed attack. In particular, we select for a starting tree
$x$ the closest neighbor $y$ with the target label $\lbl$, i.e.\ $\fun(y) = \lbl$.
Then, we use backtracing to compute the shortest script $\script$ from $x$ to $y$.
This script is guaranteed to change the label at some point.
We then apply binary search to identify the shortest prefix of $\script$ which still
changes the label (refer to Figure~\ref{fig:adv}).
\ifboolexpr{bool {arxiv}}{Refer to Algorithm~\ref{alg:targeted} for the details of the algorithm.

\begin{algorithm}
\caption{A targeted adversarial edit algorithm which transforms the input tree $x$
to move it closer to a reference tree $y$ with the desired target label $\lbl$.
The backtracing algorithm for the tree edit distance \emph{ted-backtrace} is described
in \cite{Paassen2018arxiv}.}
\label{alg:targeted}
\begin{algorithmic}[1]
\Function{targeted}{A tree $x$, a classifier $\fun$, and a
reference tree $y$ with $\fun(y) = \lbl$.}
\State $\edit_1, \ldots, \edit_\editlim \gets$ \Call{ted-backtrace}{$x$, $y$}. $lo \gets 1$. $hi \gets \editlim$.
\While{$lo < hi$}
\State $\editidx \gets \lfloor \frac{1}{2} \cdot (lo + hi) \rfloor$. $z \gets (\edit_1, \ldots, \edit_\editidx)(x)$.
\If{$\fun(z) \neq \lbl$}
\State $lo \gets \editidx + 1$.
\Else
\State $hi \gets \editidx$.
\EndIf
\EndWhile
\State \Return $(\edit_1, \ldots, \edit_{hi})(x)$
\EndFunction
\end{algorithmic}
\end{algorithm}}{}

Note that we can upper-bound the length of $\script$ by $|x| + |y|$,
because at worst we delete $x$ entirely and then insert $y$ entirely. Accordingly, our attack
finishes after at most $\effic\big( \log(|x| + |y|)\big)$ steps/queries to $\fun$.
Finally, because $y$ is the closest tree with label $\lbl$ to $x$, our attack
is guaranteed to yield a successful adversarial example if our prefix is shorter than half of
$\script$, because then $\dist(x, z) = |prefix| < \frac{1}{2} |\script| = \frac{1}{2} \dist(x, y)
= \frac{1}{2} \big(\dist(x, z) + \dist(z, y)\big)$, which implies that
$\dist(x, z) < \dist(z, y)$.
In other words, we are guaranteed to find a solution to problem~\ref{eq:adv}, in the sense
that our our label is guaranteed to change to $\lbl$, and that our solution is closest to $x$
along the shortest script $\script$ towards $y$.

\section{Experiments}

In our evaluation, we attack seven different tree classifiers on four data sets.
As outcome measures, we consider the success rate, i.e.\ the fraction of test data points
for which the attack could generate a successful adversarial example according to the definition
in Figure~\ref{fig:adv}; and the distance
ratio $\dist(z, x) / \dist(z, y)$, i.e.\ how much closer $z$ is to $x$ compared to other points $y$
with the same label as $z$. To avoid excessive computation times, we abort random adversarial
attacks that have not succeeded after $100$ tree edits. Accordingly, the distance ratio is not available
for random attacks that have been aborted, yielding some n.a.\ entries in our results (Table~\ref{tab:results}).

Our experimental hypotheses are that backtracing attacks succeed more often than random attacks
due to their targeted nature (H1), but that random attacks have lower distance ratios (H2), because
they have a larger search space from which to select close adversarials.

\paragraph{Datasets:} We perform our evaluation on four tree classification data sets from
\cite{Gallicchio2013,Paassen2018ICML}, in particular \emph{MiniPalindrome} and \emph{Sorting} as data sets
of Java programs, as well as \emph{Cystic} and \emph{Leukemia} from the biomedical domain.
The number of trees in each data set are $48$, $64$, $160$, and $442$ respectively. The
latter three data sets are (imbalanced) binary classification problems, the first is a
six-class problem.
We perform all experiments in a crossvalidation with $6$, $8$, $10$, and $10$
folds for the respective data sets, following the protocol of \cite{Paassen2018ICML}.

\paragraph{Classifiers:} On each data set, we train seven different classifiers,
namely five support vector machines (SVM) with different kernels and two recursive
neural network types. As the first two kernels, we consider the double centering kernel
(\emph{linear}; \cite{Gisbrecht2015})
based on the tree edit distance, and the radial basis function kernel (\emph{RBF})
$\kernel(x, y) = \exp\big(-\frac{1}{2} \cdot \dist(x, y)^2 / \sigma^2\big)$,
for which we optimize the bandwidth parameter $\sigma \in \R^+$ in a nested crossvalidation
in the range $\{0.5, 1, 2\} \cdot \bar \dist$, where $\bar \dist$ is the average
tree edit distance in the data set.
We ensure positive semi-definiteness for these kernels via the clip eigenvalue correction \cite{Gisbrecht2015}.
Further, we consider three tree kernels, namely the subtree kernel (\emph{ST}), which counts the number of shared proper subtrees,
the subset tree kernel (\emph{SST}), which counts the number of shared subset trees, and the partial tree kernel (\emph{PT}),
which counts the number of shared partial trees \cite{Aiolli2011}. All three kernels have a decay hyper-parameter
$\lambda$, which regulates the influence of larger subtrees. We optimize this hyper-parameter in a nested crossvalidation
for each kernel in the range $\{0.001, 0.01, 0.1\}$. For all SVM instances, we also optimized the
regularization hyper-parameter $C$ in the range $\{0.1, 1, 10, 100\}$.

As neural network variations, we first consider recursive neural networks (\emph{Rec}; \cite{Sperduti1997}), which map
a tree $x(\tree_1, \ldots, \tree_m)$ to a vector by means of the recursive function
$G(x(\tree_1, \ldots, \tree_m)) := \mathrm{sigm}\big(\mat{W}^x \cdot \sum_{i=1}^m G(\tree_i) + \vec b^x\big)$,
where $\mathrm{sigm}(a) := 1 / (1 + \exp(-a))$ is the logistic function and $\mat{W}^x \in \R^{\dims \times \dims}$ as well
as $\vec b^x \in \R^\dims$ for all $x \in \alphabet$ are the parameters of the model.
We classify a tree by means of another linear layer with one output for each of the $\nlabels$ classes,
i.e.\ $\fun(\tree) := \argmax_\lbl [\mat{V} \cdot G(\tree) + \vec c]_\lbl$, where $\mat{V} \in \R^{\nlabels \times \dims}$
and $\vec c \in \R^\nlabels$ are parameters of the model and $[\vec v]_\lbl$ denotes
the $\lbl$th entry of vector $\vec v$. We trained the network using the crossentropy loss
and Adam\ifboolexpr{bool {arxiv}}{\citep{Kingma2015}}{} as optimizer until the training loss dropped below $0.01$.
Note that the number of embedding dimensions $\dims$ is a hyper-parameter of the model,
which we fixed here to $\dims = 10$ as this was sufficient to achieve the desired training loss.
Finally, we consider tree echo state networks (\emph{TES}; \cite{Gallicchio2013}), which have the same architecture as
recursive neural networks, but where the recursive weight matrices $\mat{W}^x \in \R^{\dims \times \dims}$
and the bias vectors $\vec b^x \in \R^\dims$ remain untrained after random initialization.
Only the output parameters $\mat{V}$ and $\vec c$ are trained via simple linear regression.
The scaling of the recursive weight matrices and $\dims$ are hyper-parameters of the model, which
we optimized in a nested crossvalidation via grid search in the ranges $\{0.7, 0.9, 1, 1.5, 2\}$ and $\{10, 50, 100\}$
respectively. 

As implementations, we use the scikit-learn version of SVM, the \emph{edist} package for
the tree edit distance and its backtracing\footnote{\url{https://gitlab.ub.uni-bielefeld.de/bpaassen/python-edit-distances}},
the ptk toolbox\footnote{\url{http://joedsm.altervista.org/pythontreekernels.htm}} for the ST, SST, and PT
kernels \cite{Aiolli2011}, a custom implementation of recursive neural networks using pytorch\ifboolexpr{bool {arxiv}}{\citep{pytorch2017}}{},
and a custom implementation of tree echo state networks%
\footnote{All implementations and experiments are available at \url{https://gitlab.ub.uni-bielefeld.de/bpaassen/adversarial-edit-attacks}}.
We perform all experiments on a consumer grade laptop with an Intel i7 CPU.

\paragraph{Results and Discussion:} Table~\ref{tab:results} displays the mean classification
error $\pm$ standard deviation in crossvalidation, as well as the success rates and the
distance ratios for random attacks and backtracing attacks for all data sets and all classifiers.

\begin{table}
\caption{The unattacked classification accuracy (higher is better), attack success rate (higher is better),
and distance ratio $\dist(z, x) / \dist(z, y)$ between the adversarial example $z$, the original
point $x$, and the closest point $y$ to $z$ with the same label (lower is better) for all classifiers
and all data sets. Classifiers and data sets are listed as rows, attack schemes as columns.
All values are averaged across crossvalidation folds and listed $\pm$ standard deviation.
The highest success rate and lowest distance ratio in each column is highlighted
via bold print. If all attacks failed, results are listed as n.a.}
\label{tab:results}
\begin{center}
\begin{tabular}{lccccc}
& no attack & \multicolumn{2}{c}{random} & \multicolumn{2}{c}{backtracing}\\
classifier & accuracy & success rate & dist.\ ratio & success rate & dist.\ ratio\\
\cmidrule(lr){1-1} \cmidrule(lr){2-2} \cmidrule(lr){3-4} \cmidrule(lr){5-6}
& \multicolumn{5}{c}{MiniPalindrome} \\
\cmidrule(r){1-1} \cmidrule(lr){2-2} \cmidrule(lr){3-4} \cmidrule(lr){5-6}
linear & $0.96 \pm 0.06$ & $0.09 \pm 0.09$ & $\bm{0.24 \pm 0.07}$ & $\bm{0.52 \pm 0.15}$ & $2.68 \pm 3.54$ \\
RBF & $1.00 \pm 0.00$ & $0.06 \pm 0.06$ & $\bm{0.27 \pm 0.21}$ & $\bm{0.52 \pm 0.17}$ & $1.44 \pm 0.51$ \\
ST & $0.88 \pm 0.07$ & $\bm{0.86 \pm 0.08}$ & $\bm{0.29 \pm 0.05}$ & $0.72 \pm 0.10$ & $0.93 \pm 0.15$ \\
SST & $0.96 \pm 0.06$ & $\bm{0.78 \pm 0.15}$ & $\bm{0.36 \pm 0.08}$ & $0.54 \pm 0.11$ & $1.91 \pm 1.19$ \\
PT & $0.96 \pm 0.06$ & $\bm{0.80 \pm 0.07}$ & $\bm{0.35 \pm 0.10}$ & $0.54 \pm 0.11$ & $1.91 \pm 1.19$ \\
Rec & $0.85 \pm 0.13$ & $0.72 \pm 0.14$ & $\bm{0.17 \pm 0.05}$ & $\bm{0.79 \pm 0.08}$ & $1.26 \pm 0.41$ \\
TES & $0.92 \pm 0.06$ & $\bm{0.95 \pm 0.07}$ & $\bm{0.08 \pm 0.03}$ & $0.71 \pm 0.09$ & $1.57 \pm 0.81$ \\
\cmidrule(lr){1-1} \cmidrule(lr){2-2} \cmidrule(lr){3-4} \cmidrule(lr){5-6}
& \multicolumn{5}{c}{Sorting} \\
\cmidrule(r){1-1} \cmidrule(lr){2-2} \cmidrule(lr){3-4} \cmidrule(lr){5-6}
linear & $0.94 \pm 0.06$ & $0.02 \pm 0.04$ & $\bm{0.86 \pm 0.00}$ & $\bm{0.44 \pm 0.16}$ & $1.55 \pm 0.49$ \\
RBF & $0.94 \pm 0.06$ & $0.18 \pm 0.14$ & $\bm{0.57 \pm 0.07}$ & $\bm{0.42 \pm 0.16}$ & $1.64 \pm 0.47$ \\
ST & $0.81 \pm 0.16$ & $\bm{0.65 \pm 0.09}$ & $\bm{0.20 \pm 0.05}$ & $0.61 \pm 0.17$ & $3.01 \pm 1.91$ \\
SST & $0.89 \pm 0.10$ & $0.42 \pm 0.17$ & $\bm{0.50 \pm 0.14}$ & $\bm{0.49 \pm 0.17}$ & $1.67 \pm 0.52$ \\
PT & $0.88 \pm 0.12$ & $0.42 \pm 0.14$ & $\bm{0.52 \pm 0.17}$ & $\bm{0.50 \pm 0.14}$ & $1.69 \pm 0.87$ \\
Rec & $0.87 \pm 0.01$ & $\bm{0.64 \pm 0.20}$ & $\bm{0.44 \pm 0.07}$ & $0.26 \pm 0.17$ & $1.87 \pm 0.59$ \\
TES & $0.70 \pm 0.15$ & $\bm{0.84 \pm 0.11}$ & $\bm{0.21 \pm 0.08}$ & $0.20 \pm 0.16$ & $2.40 \pm 0.88$ \\
\cmidrule(lr){1-1} \cmidrule(lr){2-2} \cmidrule(lr){3-4} \cmidrule(lr){5-6}
& \multicolumn{5}{c}{Cystic} \\
\cmidrule(r){1-1} \cmidrule(lr){2-2} \cmidrule(lr){3-4} \cmidrule(lr){5-6}
linear & $0.72 \pm 0.09$ & $0.00 \pm 0.00$ & n.a.\  & $\bm{0.14 \pm 0.07}$ & $\bm{1.71 \pm 0.65}$ \\
RBF & $0.74 \pm 0.09$ & $0.00 \pm 0.00$ & n.a.\  & $\bm{0.22 \pm 0.13}$ & $\bm{1.68 \pm 0.56}$ \\
ST & $0.75 \pm 0.10$ & $0.00 \pm 0.00$ & n.a.\  & $\bm{0.49 \pm 0.23}$ & $\bm{0.86 \pm 0.24}$ \\
SST & $0.72 \pm 0.09$ & $0.00 \pm 0.00$ & n.a.\  & $\bm{0.34 \pm 0.16}$ & $\bm{1.25 \pm 0.32}$ \\
PT & $0.74 \pm 0.08$ & $0.00 \pm 0.00$ & n.a.\  & $\bm{0.35 \pm 0.13}$ & $\bm{1.26 \pm 0.44}$ \\
Rec & $0.76 \pm 0.11$ & $\bm{0.46 \pm 0.14}$ & $\bm{0.77 \pm 0.09}$ & $0.33 \pm 0.10$ & $1.45 \pm 1.28$ \\
TES & $0.71 \pm 0.11$ & $\bm{0.63 \pm 0.11}$ & $\bm{0.62 \pm 0.11}$ & $0.36 \pm 0.17$ & $1.23 \pm 0.28$ \\
\cmidrule(lr){1-1} \cmidrule(lr){2-2} \cmidrule(lr){3-4} \cmidrule(lr){5-6}
& \multicolumn{5}{c}{Leukemia} \\
\cmidrule(r){1-1} \cmidrule(lr){2-2} \cmidrule(lr){3-4} \cmidrule(lr){5-6}
linear & $0.92 \pm 0.04$ & $0.00 \pm 0.00$ & n.a.\  & $\bm{0.27 \pm 0.17}$ & $\bm{3.20 \pm 1.77}$ \\
RBF & $0.95 \pm 0.03$ & $0.00 \pm 0.00$ & n.a.\  & $\bm{0.20 \pm 0.08}$ & $\bm{2.88 \pm 1.65}$ \\
ST & $0.92 \pm 0.03$ & $0.00 \pm 0.00$ & n.a.\  & $\bm{0.21 \pm 0.09}$ & $\bm{2.64 \pm 0.51}$ \\
SST & $0.95 \pm 0.03$ & $0.00 \pm 0.00$ & n.a.\  & $\bm{0.19 \pm 0.10}$ & $\bm{2.57 \pm 0.65}$ \\
PT & $0.95 \pm 0.02$ & $0.00 \pm 0.00$ & n.a.\  & $\bm{0.20 \pm 0.10}$ & $\bm{2.54 \pm 0.56}$ \\
Rec & $0.93 \pm 0.03$ & $\bm{0.41 \pm 0.07}$ & $\bm{0.73 \pm 0.07}$ & $0.24 \pm 0.10$ & $2.43 \pm 0.64$ \\
TES & $0.88 \pm 0.02$ & $\bm{0.69 \pm 0.11}$ & $\bm{0.53 \pm 0.04}$ & $0.36 \pm 0.16$ & $2.49 \pm 1.11$ \\
\end{tabular}
\end{center}
\end{table}

We evaluate our results statistically by aggregating all crossvalidation folds across data sets
and comparing success rates and distance rations between  in a a one-sided Wilcoxon sign-rank
test with Bonferroni correction. We observe that backtracing attacks have higher success
rates for the linear and RBF kernel SVM ($p < 10^{-5}$), slightly higher rates for the
ST and SST kernels ($p < 0.05$), indistinguishable success for the PT kernel, and
lower success rates for the recursive and tree echo state networks ($p < 0.01$).
This generally supports our hypothesis that backtracing attacks have higher success rates (H1),
except for both neural network models. This is especially pronounced for Cystic and Leukemia
data sets, where random attacks against SVM models always failed.

Regarding H2, we observe that random attacks achieve lower
distance ratios for the ST, SST, and PT kernels ($p < 0.01$), and much lower ratios
for recursive neural nets and tree echo state nets ($p < 10^{-5}$). For the linear
and RBF kernel, the distance ratios are statistically indistinguishable. This supports H2.

\section{Conclusion}

In this contribution, we have introduced a novel adversarial attack strategy for
tree data based on tree edits in one random and one backtracing variation.
We observe that backtracing attacks achieve more consistent and reliable success across
data sets and classifiers compared to the random baseline. Only for recursive neural networks are
random attacks more successful. We also observe that the search space
for backtracing attacks may be too constrained because random attacks generally find adversarials
that are closer to the original sample. Future research could therefore consider alternative search
spaces, e.g.\ based on semantic considerations. Most importantly, our
research highlights the need for defense mechanisms against adversarial attacks for
tree classifiers, especially neural network models.

\bibliography{literature} 
\bibliographystyle{plainnat}

\end{document}